# Optical Linear Systems Framework for Event Sensing and Computational Neuromorphic Imaging

Nimrod Kruger*, Nicholas Owen Ralph*, Gregory Cohen*, Paul Hurley†*
*International Centre for Neuromorphic Systems, The MARCS Institute, Western Sydney University
†Centre for Research in Mathematics and Data Science, Western Sydney University

*Abstract*—Event vision sensors (neuromorphic cameras) output sparse, asynchronous ON/OFF events triggered by log-intensity threshold crossings, enabling microsecond-scale sensing with high dynamic range and low data bandwidth. As a nonlinear system, this event representation does not readily integrate with the linear forward models that underpin most computational imaging and optical system design. We present a physics-grounded processing pipeline that maps event streams to estimates of per-pixel log-intensity and intensity derivatives, and embeds these measurements in a dynamic linear systems model with a time-varying point spread function. This enables inverse filtering directly from event data, using frequency-domain Wiener deconvolution with a known (or parameterised) dynamic transfer function. We validate the approach in simulation for single and overlapping point sources under modulated defocus, and on real event data from a tunable-focus telescope imaging a star field, demonstrating source localisation and separability. The proposed framework provides a practical bridge between event sensing and model-based computational imaging for dynamic optical systems.

## I. Introduction

An Event Vision Sensor (EVS) approximately represent asynchronous log-intensity changes $\frac{dL}{dt}$, rather than direct absolute intensity, $I$, as is common for conventional frame-based sensors. In these complementary metal–oxide–semiconductor (CMOS) sensors, a timely event of log-intensity threshold-crossing, per pixel, is reported in relation to an internal reference level, followed by updating of this internal reference level in anticipation of the next crossing event. The ability of EVS to report changes with microsecond resolution, and with a high dynamic range compared to most image sensors, is driving many event-based applications in computer vision. However, representing these events in a physics-grounded manner, so that their information can be processed through a dynamic linear systems approach, has not been thoroughly explored in the literature. Most physics-grounded representations aim for full image-from-events reconstruction, use a scene intensity constancy assumption, or both [1]. These methods are useful for traditional imaging lenses, but don't hold for all optical systems, or are computationally expensive, and fail to exploit the relative advantage of event-sensing.

The desire to recover more scene information for highly dynamic scenes, in applications such as robotics, industrial inspection and medicine, has led to proliferation of work using EVS for computational imaging. These include EVS plenoptic cameras[2], microscopy[3]–[6], wavefront sensing[7], [8], coded aperture[9], [10], lens-less imaging[11], and more[12]–[14] - collectively coined Computational Neuromorphic Imaging (CNI)[15]. This is an emerging field that aims to expand EVS to overcome the latency-bandwidth-sensitivity bottlenecks of traditional imaging, while maintaining low system costs, energy consumption, and data bandwidth.

Computational imaging systems often use a forward operator, mapping a transfer function between scene intensity, light-field, or complex wavefront, to the sensor readout signal. Scalar-theory methods (e.g. angular spectrum and Fresnel approximation) can be used to predict sensor response to scene illumination, and also design systems for extracting specific information under various assumptions and priors[16]. With event-stream's unique data structure of binary threshold crossing response to *log*-intensity and asynchronous sampling mode, the transfer function definition and the information we put through the system is not directly obtained. Consequently, current CNI designs often reuse optical methods proven for frame-based imaging, leading to reduced performance. Alternatively, successful data-driven approaches will be hard-pressed to extrapolate, missing model-based parametrisation.

We propose a refreshed view on log-intensity change events, in the context of linear optical systems, to enable a path for CNI models and methods. This will expand the range of design possibilities for machine vision systems, microscopy, and advanced optical sensors. The following components are presented:

1) *Log-intensity derivative revisited:* we formalise a pixel's log-intensity threshold-crossing events stream to an estimate of log-intensity derivatives, intensity, and intensity derivatives via pixel event interpolations.
2) *Dynamic imaging system model:* a linear forward model for optical signals transferred through a dynamic system, represented by the operator $T(t)$, and projected onto a sensor plane. The expected intensity and intensity changes are related to the information encoded in the event-stream.
3) *Scene-informed spatial-temporal filter:* an example spatial-temporal transfer function is used to design a filter as an inverse operator on event-data to extract scene estimation.

The combined event-interpolation, known optical transfer functions (with controlled system parameters), and scene priors, enables a direct path from event-stream to scene inference. We demonstrate this path on a simulated optical model of imaging system with modulated focal length, and on real-world EVS data of a star field taken with a tunable lens imaging system. On each, the forward operator is used for source estimation using deconvolution on event-stream data.

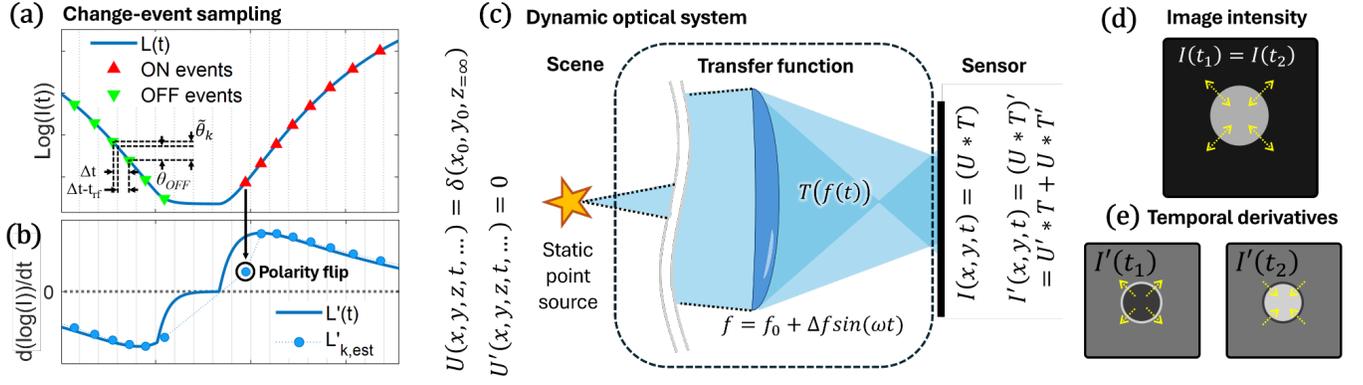

Fig. 1: Threshold crossing events for dynamic optical systems. (a) The event-pixel typically reports signal threshold crossing of log-intensity, with a set refractory period delay. (b) These are used to estimate the signal log-intensity derivative per event. A simple linear interpolation estimation works well for consecutive similar polarity events, but fails on polarity switching events. (c) A dynamic linear systems description of an optical system, $T$, projecting an input field $U$ to an intensity map $I$. The product rule describes the temporal derivative of these functions. Such a system can dynamically model a point source projection through electrically tunable thin lens. A depiction of the intensity map will present a (d) spot changing its size, while its (e) derivative for expansion at $t_1$, or contraction at $t_2$, ideally hold a uniform inter-spot value with an outer ring with opposite sign.

## II. METHODS

### A. Notation and Pixel Level Derivative Estimation

EVS's react to change in log-brightness by reporting either ON or OFF events when the pixel reference voltage crosses predefined $\theta_{ON}$ or $\theta_{OFF}$ thresholds respectively, and then reset the pixel reference voltage after a pre-defined refractory period, $t_{rf}$. The pixel voltage generally follows a log function of the impinging photon-flux intensity. An event $e_k \equiv (\mathrm{x}_k, t_k, p_k)$ represents such a change at pixel $\mathrm{x}_k = (x_k, y_k)$ at an instance $t_k$, where $p_k = 1$ or $-1$ for log-intensity increase by $\theta_{ON}$ or decrease by $\theta_{OFF}$ respectively - assuming no latency and no noise events. Figure 1(a) shows an example log-intensity signal for a single pixel, with an overlay of binary ON and OFF events where pixel voltage crosses said threshold. Specifically, for small thresholds and limited bandwidth signals, the relation between event $k$ to the log-intensity derivative can be approximated by

$$L'_k \approx \frac{\log(I_k) - \log(I_{k-1})}{\Delta t_k} \equiv \frac{L_k - L_{k-1}}{\Delta t_k}, \quad (1)$$

where $I_k$ is the signal intensity at the time of the an event, and $\Delta t_k$ is the time difference between the two, also called the inter-spike intervals. Throughout the text we use the dash ($'$) to mark *temporal* derivatives of various functions. We note the refractory period, $t_{rf}$, causes the log-intensity gap between events to be essentially different than set $\theta_{ON}$ and $\theta_{OFF}$. As shown in Figure 1(b), relating the derivative to the contrast threshold can be approximate by a linear interpolation:

$$L'_k \approx \frac{\theta(p_k)}{\Delta t_k - t_{rf}}$$
$$L_k \approx L_{k-1} + L'_k \Delta t_k \approx L_{k-1} + \frac{\theta(p_k)\Delta t_k}{\Delta t_k - t_{rf}} \equiv L_{k-1} + \tilde{\theta}_k \quad (2)$$

where $\theta(p_k)$ represents either $\theta_{ON}$ or $\theta_{OFF}$ according to event polarity value, and $\tilde{\theta}_k$ is a corrected contrast value accounting for the refractory time[17]. This approximation holds only for event couples with identical polarity (ON after ON, OFF after OFF), and extending this notion to events with polarity flipping will require additional considerations. We choose to linear interpolate the values over the last two events, and generalise the corrected contrast to be a function $\theta(p_{k-1}, p_k)$. Using this approach, we can add a multiplier for polarity flipping events, though this providing only limited improvement to the estimated value. Last, using logarithmic differentiation, we find an estimate to the intensity derivative

$$L'_k = \frac{I'_k}{I_k} \to I'_k = I_k L'_k \approx \exp\left(L_{k-1} + \tilde{\theta}_k\right)\frac{\theta(p_{k-1}, p_k)}{\Delta t_k - t_{rf}}. \quad (3)$$

For our assumed limited bandwidth signals, the intensity derivative estimated the slope *between* two events, and we make a slight change and write

$$I'_k = \left(\frac{I_{k-1} + I_k}{2}\right)L'_k. \quad (4)$$

Here all values are, in fact, *estimations* and written as true signal values only for convenience.

Expressing per-event intensity derivative, $I'_k$, hold several merits. First, as mentioned, it is possible to express the event-stream by a linear system operator on the input scene (or scene derivative). Second, we can use knowledge of either scene or system *dynamics* to better infer scene information. In addition, as events represent timely *changes* in intensity, downstream computation directly processing the magnitude of these changes can enjoy both the sparse asynchronous nature of EVS and accurate performance. Specifically, we can compute and assign a *value* to each event that is related to a physical quantity (the rate change of photon-flux).

## B. Single Pixel Signal Estimation

Considering equation 4, we perform a rough evaluation of the quality of this reconstruction, and its characteristics. We acknowledge the approach represents many first-order approximations, and a full statistical modelling of event inter-spike intervals [18], [19] and interpolation methods for high bandwidth signals are required for accurate interpolation-based reconstruction.

An ideal event-pixel will report threshold crossing without delay, then reset to precisely the value of the signal after the refractory period, and no "noise" events will be reported. For a log-intensity signal with ideal event sampling, as shown in Figure. 2(a), we can reach relative good approximation of the original signal as in Figure. 2(b - blue dots). The log-intensity derivatives, Figure. 2(c - blue dots), and intensity derivatives, Figure. 2(d - blue dots) also mostly follow their ground truth signals, with the exception of polarity flipping events (as highlighted in Fig. 1(b)).

However, real event data will naturally include noise events of many shapes and forms (Fig. 2($a_2$)). These accumulate to an intensity reconstruction error, exacerbated by the log-amplifying nature of EVS, as shown in Figure. 2($b_2$). The log-derivative estimation is hardly affected by this process (Fig. 2($c_2$)), resulting in only a change in the proportion of the intensity derivative estimation for each event (Fig. 2($d_2$)). Specific noise events will create a corresponding noisy derivative measurement, but with limited downstream repercussions.

## C. Dynamic Linear System for Event Data

Figure 1(c) depicts the relation between a scene $U$, a dynamic space-invariant linear system expressed as $T$, and the resulting projection $I$. For a system with such a $T$ function, and expressing these as 2D matrices, we can express the projection as

$$I = T *_s U, \quad (5)$$

where $*_s$ is the 2D-*spatial* convolution operation. The dynamics of this projection will follow

$$I' = \frac{\partial (T *_s U)}{\partial t} = \frac{\partial T}{\partial t} *_s U + T *_s \frac{\partial U}{\partial t}. \quad (6)$$

For simplicity, we consider only static scenes with a dynamic projection systems, such that $I' = T' *_s U$, and the general case is left for the discussion. Considering a direct raw deconvolution approach a scene can be reconstructed from the event-stream by expressing the inverse filter of the system:

$$U_{est} = \mathcal{F}_i \left\{ \frac{\mathcal{F}\{I'\}}{\mathcal{F}\{T'\}} \right\} = \mathcal{F}_i \left\{ \frac{\hat{I}'(\bar{\mathbf{k}})}{\hat{T}'(\bar{\mathbf{k}})} \right\}, \quad (7)$$

where $\mathcal{F}$ represents a Fourier transform and $\mathcal{F}_i$ the corresponding inverse Fourier transform, and $\hat{I}'(\bar{\mathbf{k}})$ and $\hat{T}'(\bar{\mathbf{k}})$ represent the derivative image and optical transfer function in the frequency space $\bar{\mathbf{k}}$. This method is limited to frequencies within a finite range where $\hat{T}'(\bar{\mathbf{k}}) > 0$, to avoid isolated poles, and neglects the effect of noise amplified by deconvolution. To address these, we consider the Wiener deconvolution[16] by expressing

$$U_{est} = \mathcal{F}_i \left\{ \frac{\hat{T}'^{*}(\bar{\mathbf{k}}) \hat{I}'(\bar{\mathbf{k}})}{\left|\hat{T}'(\bar{\mathbf{k}})\right|^2 + \lambda(\bar{\mathbf{k}})} \right\}, \quad (8)$$

where $\hat{T}'^{*}(\bar{\mathbf{k}})$ is the complex conjugate of the deconvolution filter, and $\lambda(\bar{\mathbf{k}})$ is a heuristic regularisation, and can be replaced by the noise mean spectral density when it is known. Here we use a linear function $\lambda(\mathbf{k}_r) = \eta \mathbf{k}_r$, where $\eta$ is manually tuned.

Inverse filters in the context of image reconstruction are conventionally performed on dense 2D matrices. For simplicity, we employ a 2D Fast Fourier Transform (FFT) to analyse both event derivatives and $T'$. This is but a placeholder for event-native Fourier transforms that will be explored in future work[20], [21]. A frame is constructed choosing only the most recent derivative value per pixel, multiplied by a decay factor (according to the time since that event was reported) before performing a 2D FFT on the image.

As an example, consider imaging a point source by a dynamically varying focal length system as in Figure 1(c). Specifically, consider a system similar to that of Ralph, Maybour, Marcireau, *et al.* [22], where a stationary star field (neglecting atmospheric turbulence) is imaged using a telescope fixed liquid lens. Modulating the focal length around the sensor focal point allowed for improved noise rejection, star brightness estimation, and avoiding the meticulous chore of focusing an event camera. For now, however, we only consider an equivalent system with thin-lens approximation, monochromatic point sources, narrow field-of-view, and no additional aberration. For such a system, the projection of a point source will be a uniformly expanding and contracting spot with a constant overall intensity as shown in Figure 1(d+e), where the transfer function and its derivatives can be analytically described (see supplementary materials).

## D. Spatial-Temporal Kernels

We consider the expression $\hat{T}'(\bar{\mathbf{k}})$ in equation 7 and 8, dubbed here the deconvolution kernel. This can either be analytically calculated from a projection model, estimated using simulation, or measured by creating an impulse response. Measured impulse responses could become useful for encapsulating both system dynamics and sensor stochastic nature, but require dedicated measuring and calibration procedures. We note, for instance, that the interpolation method leans heavily on knowledge of the contrast threshold, and non-uniformity calibrations[23], [24] may prove critical. Analytical calculations, alternatively, require deriving the PSF of a system, and its temporal derivative, before performing a 2D Fourier transform.

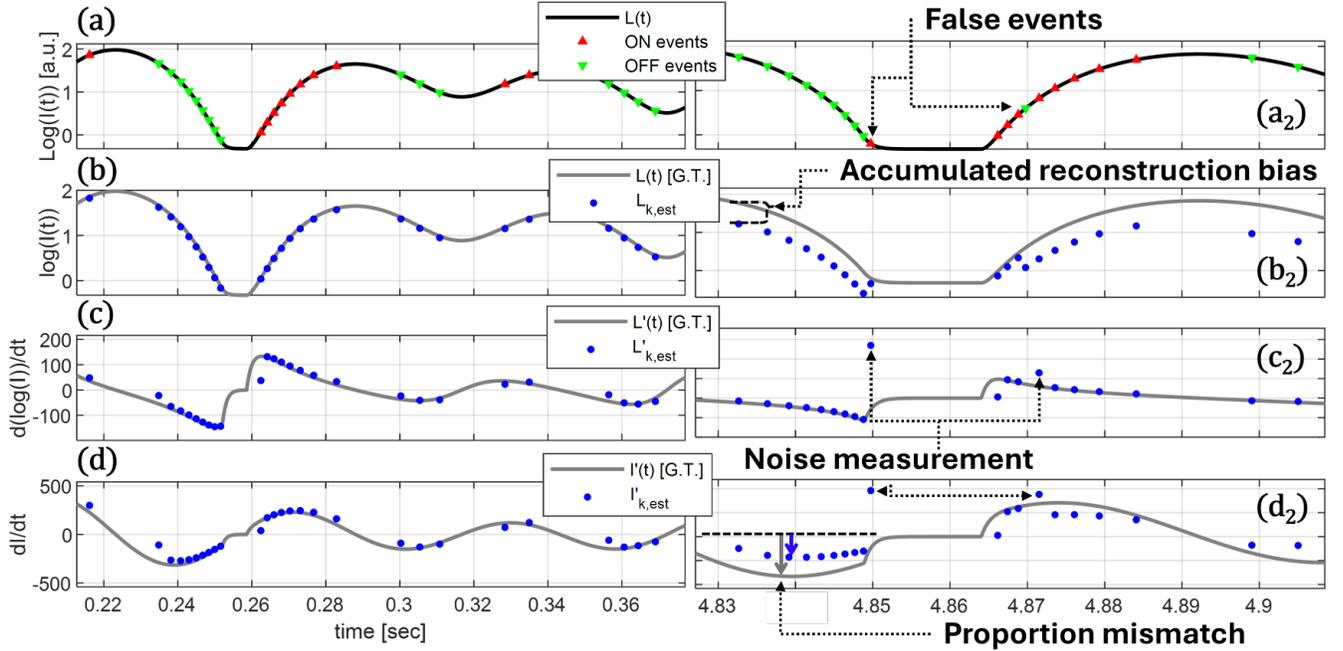

Fig. 2: Pixel signal inference illustration. (a) A reference signal is measured by reporting ON and OFF events for each crossing of a corresponding threshold in its log-intensity profile. (b) A signal log-intensity estimation can be modelled by accumulating threshold according to the ON and OFF event count, (b$_2$) prone to noise related reconstruction biases. (c) A first order estimate of the log-derivative can be used to (d) estimate each event's intensity derivative according to the product of the log-intensity derivative and intensity estimation. (d$_2$) Noise events will lead to localised derivative miscalculations, and reconstruction bias will lead to mismatch in the proportion of the estimated derivatives.

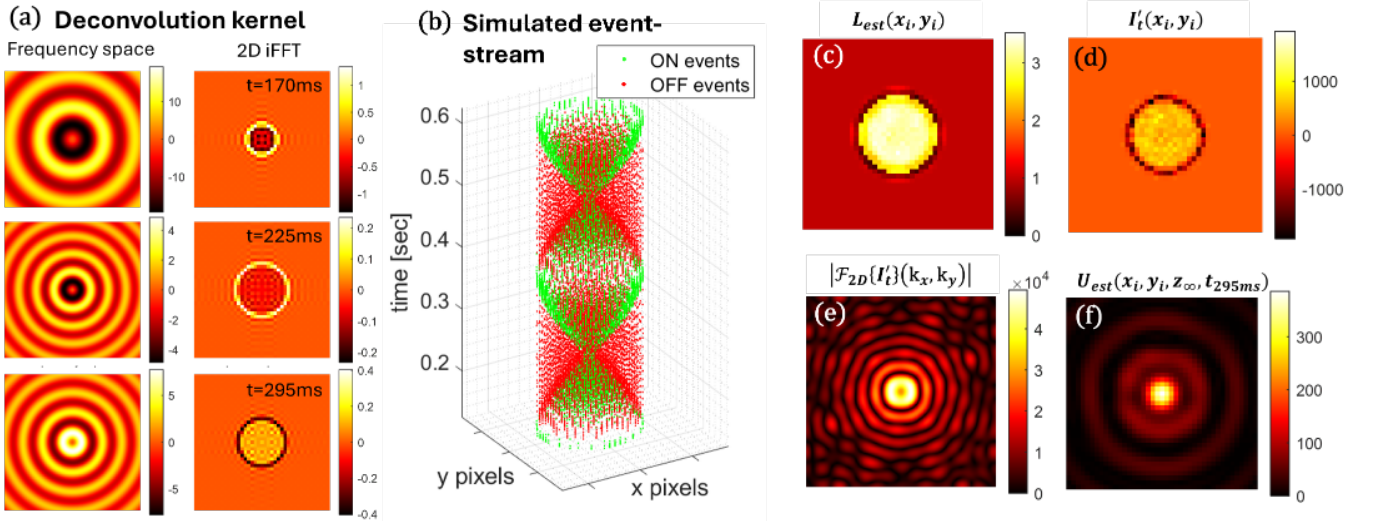

Fig. 3: Estimating point source via deconvolution of event streams. (a) A dynamic deconvolution kernel for a tunable lens system is defined analytically - presented here in both frequency domain and in pixel-space for different moment in time. (b) A simulation generates an event-stream for our point-source through a modulated focus optical model. The algorithm calculates the (c) log-intensity, (d) event-based intensity derivative, and the (e) Fourier domain of that image - and then (f) a raw-deconvolution estimation the source. There are presented here for t=295 ms after simulation onset.

Doing so for a tunable thin lens system we can express the deconvolution kernel as

$$\hat{T}'(\mathbf{k}_r, t) = 2\pi R(t) \cdot \left( g(t) \frac{J_1(R\mathbf{k}_r)}{\mathbf{k}_r} + h(t) J_0(R\mathbf{k}_r) \right), \quad (9)$$

where $R(t)$ represents the momentary spot radius (a function of the effective focal length and other system parameters), $J_n(x)$ is the Bessel function of order $n$, and $\mathbf{k}_r$ is the radial frequency, translated to sensor's frequency domain by $\mathbf{k}_r = \sqrt{\mathbf{k}_x^2 + \mathbf{k}_y^2}$. Here $g(t)$ and $h(t)$ are the inter-spot intensity profile, and the outer ring impulse-function respectively (these are evolving functions representing the change in spot shape, and are detailed in the supplementary material). Figure 3(a) shows the images both in frequency space, and their spatial representation on a 2D pixel matrix at different points in time - two during spot expansion stage, and one on contraction. The spatial-domain images shows the expected outer ring with high derivative values (representing transition from spot illumination to background illumination or vice versa), and the internal derivative value representing increase (or decrease) as spots concentrate illumination on fewer pixels.

However, not all systems can be as easily modelled. Combined event and optical simulators[25]–[28] to estimate the deconvolution kernel present a useful middle ground. While generalised event simulators are not yet mature or computationally efficient, their optical counterpart are. This allows for developing first order deconvolution kernels according to a system's design.

We also comment on an additional avenue to implement spatial-temporal priors in our framework - introduced even earlier in the computation. When accumulating the intensity estimation map (equation 4), one might choose to expand the pixel-level estimation to considering the local intensity map - and introduce priors related to the scene and pixel behaviour. To clarify this notion by way of example, often in EVS data we experience hyper-sensitive pixels (or even "hot pixels" that fire constantly). While these are often treated by a a pre-processing filter stage, we implement a different approach. We dictate accumulated intensity estimation, $I(x_k, y_k)$, to also be mostly by *neighbouring* pixel events, rather solely from the events it reports. If no such events arrive, the intensity estimation hardly changes, and the derivative estimation remains small. However, if neighbouring pixel fire just as often (as will happen for true changes), the local pixel cluster intensity will rise in unison. when an event occurs at $(x_k, y_k)$ we add $\exp(L_k) \cdot ker_{NN}$ to the intensity map centred at $(x_k, y_k)$, where

$$ker_{NN} \equiv \begin{bmatrix} 0.065 & 0.178 & 0.065 \\ 0.178 & 0.024 & 0.178 \\ 0.065 & 0.178 & 0.065 \end{bmatrix}. \quad (10)$$

This is successfully used on real-world EVS data, and merits future work of theoretical analysis of this mode.

### E. Scene Estimation Algorithm

These derivations can essentially be performed per-event. However, as we are using the frame-based 2D FFT placeholder, repeating this for each new event is computationally expensive. Therefore we estimate the sensor intensity derivative, $I'_t(x_i, y_i)$, at uniformly separated time samples $t$, similar to the description above - with several differences. **a)** We accumulate events within a time step to update the intensity estimation, using a continuous-time filter as used in Scheerlinck, Barnes, and Mahony [29], and starting off with an initial guess. The guess is based, when possible, on the original input to the event simulator, or using the overall event count from the recording. **b)** The image derivative is updated using only the most recent events for each pixel within a time step, and an additional decay factor of $\exp(-(t - t_k)/\tau)$. **c)** For each time-step we perform a 2D FFT operation on the derivative image $I'_t(x_i, y_i)$ to get $\hat{I}'(\mathbf{k}_x, \mathbf{k}_y)$. Zero padding of the original image was used to smoother frequency-space images.

## III. RESULTS

### A. Deconvolution of Simulated Data

For our proof of concept we simulate an event-stream using [22], as presented in Figure. 3(b). The image frames provided to the simulation are of an expanding and contracting spot according to a sinusoidally modulated tunable lens model. Optical parameters of the model are chosen to be similar to those of our liquid-lens on a telescope imaging of a star field. Pixel parameters were mostly chosen to fit the EVS Prophesee Gen4 device values (size, dark-current, etc.), while others were chosen for developing this framework - such as deliberately introducing non-uniformity across pixel contrast threshold values (chosen $0.2 \pm 0.02$), but no extra noise events. We note that uniform contrast thresholds across all pixels led to symmetries in the event projections - creating misleading, albeit beautiful, frequency images.

Following the interpolation of events, as described in II-E, we produce log-intensity estimation frame (Fig. 3(c)) and intensity derivative estimation image (Fig. 3(d)). The 2D-FFT is calculate (Fig. 3(e)), and used to estimate the point source image at infinity according to equation 8 with a uniform mean spectral density value, and a low-pass filter imposed on the resulting spectral image. A typical image resulting from this process is seen in Figure 3(f). In this example we displayed not the best nor the worst result, and a full video of these 4 images (Fig.3(c,d,e,f)) for multiple cycles of the focal length modulation is provided as supplementary material. As expected, during periods with limited dynamics (the spot is at either maximum or minimum, and the dynamic optical transfer function becomes a null operator) and many polarity flips, the source estimation diverges - both in shape and intensity estimation.

To further express the power of this tool, we test simulated data presenting two point-sources at close vicinity - such that the out-of-focus imaging system will lead to overlapping events (events generated at pixels that are affected by both sources intensities) over prolonged periods of the lens modulation. We choose one source to be double the intensity of the other, and position the points at various distances from each other. In Figure 4(a) the event-stream for points positioned 5.6 pixels apart is portrayed, with the a corresponding momentary intensity derivative estimation (Fig. 4(b)), and the resulting source estimation shown in Figure 4(c). In this step we implemented a estimated linear rise of the mean spectral density of the noise into the Wiener deconvolution.

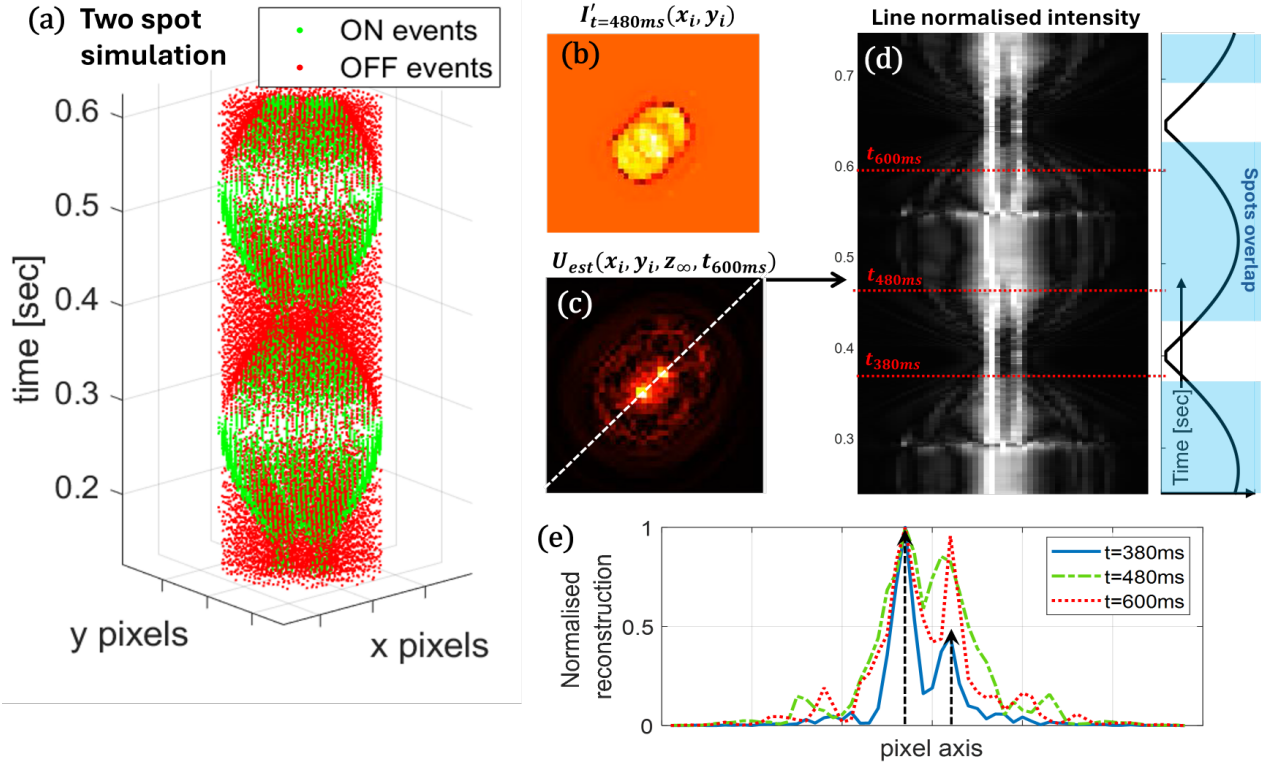

Fig. 4: Deconvolution of multiple sources. (a) A simulated event-stream of two point-sources, one with double the brightness relative to the other, and distanced by 5.6 pixels, was generated. (b) The intensity derivative was calculated and the (c) source inference estimated. (d) The normalised intensity of the line crossing both source-point locations is plotted as a function of time. On the right of the image the corresponding spot radius is displayed, highlighting periods where the imaged spots overlap. (e) For select time-frames, $t = 380$ ms, 480 ms, & 600 ms, the normalised line intensity is plotted - with an addition of the expected ground truth point source intensity and position marked by dashed arrows.

Processing these inference images throughout the sequence, we extract only the average along the line where the two source point are positioned on (Fig. 4(c) - dashed white line), and plot the normalised value along the line as a function of time (Fig. 4(d)), with an addition of the expected spot radius on the right for reference. Periods where the two spots overlap on the sensor image plane are highlighted in blue. In this demonstration, we see the output spots are mostly noticeable and separable. Exceptions include the time of the expanded spot *polarity-flip* event (roughly around 300 ms & 550 ms), fitting the issues we described in our log-derivative interpolation at these instances. In addition, during the minimal spot radius (imaging in focus), there are occasional disappearance of the spots - probably due to no imaging system motion at these periods. Highlighting specific areas, we see optimal source estimation occurs when the spots are also separable on the image plane, such as at $t =380$ ms (Fig. 4(e) - blue solid line) - where the right source presents the reduced intensity fits the expected 50% intensity compared to the left source (see Fig. 4(e) black arrows for source ground truth). When spots are expanding (e.g. $t =480$ ms), or contracting (e.g. $t =600$ ms), the two sources are also clearly reconstructed - but with both reduced accuracy in relation to relative intensity and various background spatial features (Fig. 4(e) - dotted red line and dashed green line + Fig. 4(c)).

### B. Synthetic Deconvolution Kernel for Measured Event Data

In this section we test the method on data collected in [22], where events are generated by modulating the focal length of a 2800 mm focal length telescope imaging the star field 'M47' (Messier 47/NGC 2442). The telescope was equipped with Optotune El-10-30-C-VIS liquid lens system, and modulated to obtain a sinusoidal function of the focal length, dynamically projecting the star field in a similar fashion to the optical model described above. Application of the deconvolution method quickly presents the deviation from this model, foremost as the imaging of broad-spectrum image through a liquid lens system is hard pressed by the thin lens approximation. For instance, slight spherical aberrations will result in a highly non-uniform PSF when out of focus, also breaking the symmetry assumed for focusing behind and before the sensor. A realistic PSF, possibly derived using an optical simulator, would provide a spot with more energy on the periphery when the focused behind the detector, and a central-heavy energy distribution when focused before the detector. In addition, dynamic behaviour that is unrelated to the focal length modulation is present - namely atmospheric turbulence. While these are interesting to consider, we would need more modes to validate any observed wavefront shifts.

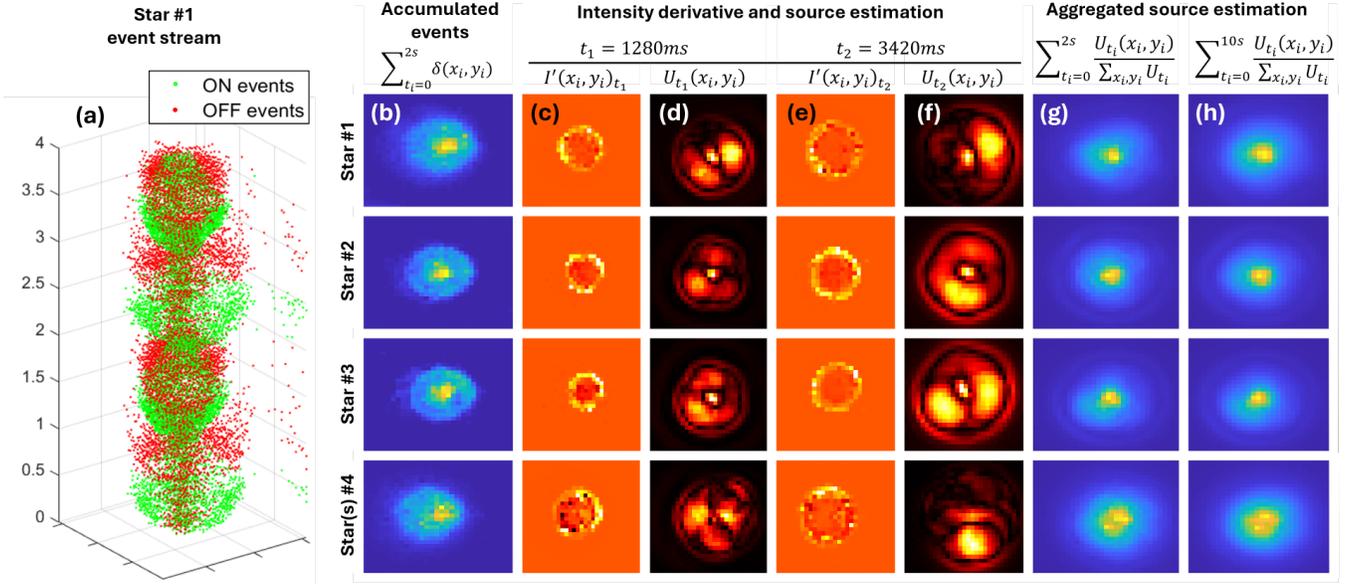

Fig. 5: Sample source estimation from measured EVS data on modulated focal length images of a star field. (a) Example event stream of a 0.5Hz modulated image of a star. Deconvolution from events on 4 different stars is performed, where for each the (b) 2 seconds accumulated raw events, (c,e) sample intensity derivative and (d,f) source estimations for two points in time, and the average normalised source estimation over (g) 1 cycle and (h) 5 cycles are presented.

Nonetheless, for extended periods within the processing cycle, we are able to obtain a clear point source inference, with additional side-lobes representing various dynamics and residual optical features. Out of the data from the M47 star field, we consider 4 sources and run deconvolution on event data from their vicinity. We choose data from the lens modulation frequency of 0.5 Hz, having a maximal number of events per cycle, and a sample event stream for the first 2 cycles is presented in Figure 5(a). The stars #1-3 are only single sources while star #4 is, in fact, two sources positioned in close proximity. In Figure 5 we show the processing results for each of these source. These include (b) the accumulated event count per pixel for one modulation cycle, (c,e) the estimated intensity derivative per pixel and (d,f) resulting source estimation for a point in time during the (c,d) first cycle, and the (e,f) second cycle - both when the focal point is behind the sensor plane (where the PSF is similar to the thin lens model PSF).

Considering the overall effect of this process, we accumulate normalised source estimation results over the duration of entire cycles. As shown in Figure 5(g), one cycle is enough to provide sharper point-source localisations compared to just raw events accumulation. In addition, for star #4, the two sources are easily separable with a distance of less than 3.5 pixels. With addition of 5 cycle averaging (Fig. 5(h)), point-sources are often better localised, with the exception of star #1 - possibly due to atmospheric turbulence effects.

The source estimations for the different single stars (#1-3) all present a clear single point, surrounded by directional lobs - representing the direction where a spot is expanding quicker to. For the star-duo #4, it is harder to attribute characteristics that contribute to the image. Closer inspection showed that information contributing to this image being a star-duo comes from all part of the processing period, but mostly from when the PSF is in focus. A full video stream of the processed EVS data, exact labels of the sample stars, and a full M47 star field reconstruction are available in the supplementary materials.

## IV. Discussion

In this work, we suggest a new framework for EVS data processing, starting the discussion on a generalised imaging approach of applications such as wavefront sensing, light-field imaging, and coded aperture. These have in common an optical projection design that doesn't follow the intuitive pin-hold camera model, or include some sort of engineered dynamics. However, wider applications will present themselves, as progress is made on theoretical and computational aspects. For one, our reconstruction and derivative estimation leaves ample room for improvement - with issues related to polarity switching, accuracy around drastic intensity jumps, and handling non-ideal pixel characteristics (such as light-dependant latency, threshold non-uniformity, and detailed noise estimation). As EVS sensor models develop, so will the underling assumptions used for event interpolation - and lead to better intensity and intensity derivative estimations. Interpolation methods using more than the last two events can be developed for this goal. That said, many successful intensity-from-events reconstruction methods are available in the literature [29]–[32] (highlighting non data-driven methods for imaging systems), and may be adapted to estimate intensity derivatives.

As for handling a task-dependant system with both scene and operator dynamics, we have several options for full scene inference. For instance, the computation would be separated to slow and fast dynamics, assuming quasi-static scene, or quasi-static operator, and calculate the inference on different timescales. Alternatively, regression would be deployed on both the static and dynamic scene outputs, minimising the sum of the projection of dynamic and static operations - though failure modes need to be carefully explored here. Additionally, the use of frame-based accumulation was only chosen to accommodate conventional 2D-FFT tools, eroding the benefits of the sparse asynchronous event representation. Replacing these with event-based Fourier transforms[20], [21], or Non-Uniform Fast Fourier Transform (NUFFT), may improve performance - and inspire applications where image reconstruction is not the goal. The example closing the results section (Fig.5), with static sources projection on the sensor affected both by the imaging system and atmospheric dynamics, suggests simultaneous event-based decoding of both scene and optical system. Thus, layers of flexibility may be added to EVS systems, including control of system parameters (such as spatial bandwidth, spectral encoding, etc.), to extract static characteristics of the scene, and detect scene dynamics with rates surpassing any conventional frame-based sensor.